\documentclass[lettersize,journal]{IEEEtran}
\usepackage{amsmath,amsfonts}
\usepackage{algorithmic}
\usepackage{algorithm}
\usepackage{array}
\usepackage[caption=false,font=normalsize,labelfont=sf,textfont=sf]{subfig}
\usepackage{textcomp}
\usepackage{stfloats}
\usepackage{url}
\usepackage{verbatim}
\usepackage{graphicx}
\usepackage{cite}
\usepackage[table]{xcolor} 
\usepackage{booktabs}
\usepackage{makecell}
\usepackage[colorlinks=true,urlcolor=magenta]{hyperref}

\begin{document}

\title{Vision-LLMs for Spatiotemporal Traffic Forecasting}

\author{Ning Yang, Hengyu Zhong, Haijun Zhang, Randall Berry
\thanks{Ning Yang is with the Institute of Automation, Chinese Academy of Sciences, Beijing, 100190, China (e-mail: ning.yang@ia.ac.cn).}%
\thanks{Hengyu Zhong is with Southwest University, Chongqing, 400715, China. This work was performed while he was an intern at the Institute of Automation, Chinese Academy of Sciences (e-mail: hengyuzhong@126.com).}%
\thanks{Haijun Zhang is with the Department of Computing and Communication Engineering, Beijing University of Science and Technology, Beijing, 100083, China (e-mail: zhanghaijun@ustb.edu.cn).}%
\thanks{Randall Berry is with the Department of Electrical and Computer Engineering, Northwestern University, Chicago, 60208, USA (e-mail: rberry@northwestern.edu).}%
\thanks{\textit{(Corresponding author: Ning Yang.)}}
}

\markboth{}%
{}


\maketitle

\begin{abstract}
Accurate spatiotemporal traffic forecasting is a critical prerequisite for proactive resource management in dense urban mobile networks. While large language models have shown promise in time series analysis, they inherently struggle to model the complex spatial dependencies of grid-based traffic data. Effectively extending large language models to this domain is challenging, as representing the vast amount of information from dense geographical grids can be inefficient and overwhelm the model's context. To address these challenges, we propose \textbf{ST-Vision-LLM}, a novel framework that reframes spatiotemporal forecasting as a vision-language fusion problem. Our approach leverages a Vision-LLM visual encoder to process historical global traffic matrices as image sequences, providing the model with a comprehensive global view to inform cell-level predictions. To overcome the inefficiency of large language models in handling numerical data, we introduce an efficient encoding scheme that represents floating-point values as single tokens via a specialized vocabulary, coupled with a two-stage numerical alignment fine-tuning process. The model is first trained with supervised fine-tuning and then further optimized for predictive accuracy using group relative policy optimization, a memory-efficient reinforcement learning method. Evaluations on real-world mobile traffic datasets demonstrate that ST-Vision-LLM outperforms existing methods by 15.6\% in long-term prediction accuracy and exceeds the best baseline by around 30\% on average in cross-domain few-shot scenarios. Our extensive experiments validate the model's strong generalization capabilities across various data-scarce environments.
\end{abstract}


\section{Introduction}
\IEEEPARstart{T}{he} ever-increasing demand for high-speed, reliable mobile connectivity in dense urban environments presents a significant challenge for network operators. Meeting this demand hinges on proactive resource management, for which accurate traffic prediction is a critical prerequisite \cite{ref_6}. The evolution of spatiotemporal sequence prediction has progressed from classical statistical methods to more advanced deep learning approaches. Early deep learning efforts utilized sequence models, such as those based on recurrent or temporal convolutional architectures, which demonstrated proficiency in capturing temporal patterns for tasks like traffic flow prediction \cite{ref_2, ref_3}. To better address data with explicit spatial structures, subsequent research integrated graph-based techniques. These graph neural network models leverage mechanisms like graph convolutions and attention to simultaneously model complex spatial and temporal dependencies \cite{ref_4, ref_21}. Although these methods have achieved promising results in their respective application scenarios, they often require meticulous architectural design tailored to specific datasets and tasks, thus exhibiting limited generalization capabilities across different scenarios \cite{ref_6}.

Inspired by the success of large-scale pre-trained models in computer vision (CV) and natural language processing (NLP), researchers have begun exploring general-purpose time series analysis models. Foundation models, particularly those based on the transformer architecture, have shown superb performance by leveraging self-attention mechanisms and sophisticated input representations to enhance long-term prediction capabilities \cite{ref_8, ref_7}. Similarly, methods that adapt Large Language Models (LLMs) for time series forecasting have emerged, harnessing the powerful representation and reasoning abilities of LLMs through techniques like text-based prompting or fine-tuning \cite{ref_9, ref_10}. However, despite their proficiency in capturing intricate temporal dependencies and patterns, these general-purpose approaches often fall short in modeling the complex spatial associations and topological structures inherent in spatiotemporal traffic data \cite{ref_11, ref_10}.

Recognizing this limitation, several recent works have attempted to bridge the gap by integrating spatial structural information with LLMs. These approaches generally involve strategies such as augmenting the input sequence with specialized spatiotemporal positional encodings \cite{ref_12}, tokenizing graph-structured neighborhoods for processing by the LLM \cite{ref_13}, and directly modifying the core attention mechanism or adding adapter modules to inject a spatial bias \cite{ref_14}. A parallel line of work uses dedicated spatiotemporal encoders to pre-process the data before instruction-tuning the LLM \cite{ref_15, ref_51}. However, these methods face a common set of challenges. They either append spatial information as linear sequences, which can be an inefficient representation that struggles with information compression, or they necessitate substantial modifications to the model's core architecture to handle spatiotemporal features \cite{ref_15, ref_51, ref_12, ref_13, ref_14}. Such intrusive approaches risk constraining the model's ability to leverage the full reasoning and generalization capabilities inherent in the pre-trained LLM.

The central challenge is thus to integrate complex spatial dependencies into an LLM framework without resorting to inefficient data representations or compromising the integrity of the pre-trained model's architecture. To overcome this, we propose the \textbf{Spatiotemporal Vision Large Language Model (ST-Vision-LLM)} framework, which reframes the spatiotemporal sequence prediction task as a vision-language fusion problem. Instead of treating spatial locations as a discrete list, ST-Vision-LLM employs a visual encoder to transform historical global traffic matrix sequences into holistic image-like representations. These visual embeddings are then integrated into the context of a Vision-LLM, enabling the model to perceive and reason about comprehensive global spatiotemporal patterns as a unified scene, thereby naturally capturing spatial relationships during cell-level predictions. To address the inefficient representation of numerical values in standard LLMs, we design a float-specific vocabulary that encodes numbers into single tokens, significantly conserving context length. We then perform a two-stage numerical alignment fine-tuning process to enable the model to comprehend and generate these new numerical tokens. With this enhanced numerical capability established, we first conduct supervised fine-tuning on spatiotemporal traffic data, followed by policy optimization using a memory-efficient reinforcement learning algorithm \cite{ref_16}, to further refine and improve prediction accuracy.

\textbf{Our main contributions include:}
\begin{enumerate}
    \item We propose the ST-Vision-LLM, a novel framework that reframes spatially correlated spatiotemporal sequence prediction as a vision-language fusion task. This approach transforms the prediction problem into a sequence generation task within a multimodal context.
    \item Our work introduces the ST-Vision-LLM architecture, which embeds global historical spatiotemporal information into the LLM context via a visual encoder. This framework also introduces an efficient numerical token encoding and an aligned fine-tuning strategy, enabling the model to perform cell-level traffic prediction based on a complete global view.
    \item The effectiveness of our approach is validated on multiple real-world mobile network spatiotemporal datasets. The results demonstrate that ST-Vision-LLM surpasses existing methods in long-term prediction accuracy and exhibits outstanding performance in few-shot and zero-shot scenarios, offering an effective solution for spatiotemporal traffic prediction in data-scarce environments.
\end{enumerate}

\section{Related Work}
\subsection{Task-Specific Learning}
Traditional spatiotemporal sequence prediction methods are often customized for specific tasks and domains, training models end-to-end on small-scale datasets. Early efforts in spatiotemporal sequence prediction began by adapting models from time series analysis. For example, the classical statistical model Autoregressive Integrated Moving Average (ARIMA) was applied to traffic flow prediction \cite{ref_18}, but its linearity struggled with complex dynamics. To address this, deep learning models were introduced. Recurrent networks like Long Short-Term Memory (LSTM) could capture non-linear temporal dependencies, with hybrid models like ResLSTM \cite{ref_19} combining it with residual and graph convolutional networks. Convolutional networks such as Temporal Convolutional Networks (TCNs) proved effective for long sequences, and architectures like ST-ResNet \cite{ref_20} utilized residual networks to model various temporal properties. However, these models primarily focused on temporal patterns, neglecting spatial correlations.

To address this spatial oversight, researchers introduced graph neural networks (GNNs). Early models such as Diffusion Convolutional Recurrent Neural Network (DCRNN) and Spatio-Temporal Graph Convolutional Networks (STGCN) \cite{ref_21} combined graph convolutions with recurrent or convolutional layers to model predefined spatial and temporal dependencies jointly. However, their reliance on static predefined graphs limited the modeling of dynamic spatial correlations, motivating adaptive adjacency matrices in models such as Graph WaveNet (GWNET) \cite{ref_22}, which learn spatial dependencies directly from data. Further variants such as Multi-Component Spatial-Temporal Graph Convolution Networks (MCSTGCN) \cite{ref_23} were proposed to capture correlations across different time periods. Attention mechanisms were also introduced to weigh different spatiotemporal relationships, as in the Graph Multi-Attention Network (GMAN) \cite{ref_46} and Attentive Crowd Flow Machine (ACFM) \cite{ref_24}, the latter using two progressive ConvLSTM units. Despite these advances, such methods are still typically designed and trained end-to-end for narrow tasks, for example Spatio-Temporal Network (STN) for mobile traffic \cite{ref_6}, and therefore often generalize poorly to diverse time series data.

\subsection{Time Series Level General Learning}
Inspired by the success of pre-training and fine-tuning in CV and NLP, recent studies have explored general time series models by pre-training transferable representations on large-scale data. Early efforts focused on improving the Transformer architecture for long-term forecasting, with models such as Informer, Autoformer, and FEDformer designed to improve efficiency and handle long dependencies. A significant conceptual shift came with Patch Time Series Transformer (PatchTST), which tokenizes time series into sub-sequence ``patches'', allowing Transformers to better capture local semantic information.

This breakthrough paved the way for leveraging the power of pre-trained LLMs. Frozen Pretrained Transformer (FPT) demonstrated that even a frozen pre-trained Transformer can achieve strong performance by only fine-tuning the head and tail. Further works like Time-LLM \cite{ref_10} and LLM4TS \cite{ref_9} explored reprogramming and fine-tuning strategies to align LLMs with time series data, unlocking powerful few-shot and zero-shot capabilities. While models like iTransformer \cite{ref_47} further refined the architecture to better capture multivariate correlations, these generalist approaches are fundamentally designed for one-dimensional (1D) sequences. They inherently lack the mechanisms to model the complex topological structures and spatial correlations present in two-dimensional (2D) grid-based spatiotemporal data. Related work also includes pre-training time series foundation models from scratch, such as TimesFM \cite{ref_48}, CHRONOS \cite{ref_49}, and MOMENT \cite{ref_50}, which train models using large-scale time series data and achieve significant prediction performance in zero-shot and few-shot scenarios. In summary, these methods primarily focus on capturing temporal dependencies and patterns in time series data but generally lack spatial modeling capabilities, making them unsuitable for scenarios requiring complex spatial analysis.

\subsection{Spatiotemporal Series Level General Learning}
Recognizing the importance of spatial correlations, recent works have combined spatial information with LLMs for spatiotemporal sequence prediction. These approaches generally follow two main strategies. One strategy involves using separate, specialized encoders to process spatial and temporal features before feeding them to an LLM. UrbanGPT \cite{ref_15}, for instance, uses a temporal convolutional encoder, while Traffic Prediction LLM (TPLLM) \cite{ref_51} employs both convolutional and graph convolutional networks. A second, more integrated strategy aims to make the LLM itself spatially aware. This is often achieved by either enriching the input with positional information, as seen in Spatial-Temporal Large Language Model (ST-LLM) \cite{ref_12}, or by designing novel tokenizers like the graph-based tokenizer in STG-LLM \cite{ref_13}. Other approaches fuse graph neural networks with LLMs, such as GCNGPT \cite{ref_12}, which combines GCN with a pre-trained Transformer, and GATGPT \cite{ref_25}, which integrates graph attention mechanisms. Some approaches, including ST-LINK \cite{ref_14}, even modify the core attention mechanism to explicitly encode spatial relationships. A recent example is Jiang et al. \cite{ref_45}, who combine LLM-based knowledge priors with dynamic correlation modeling for lane-level traffic flow prediction.

However, these methods are primarily designed for data with discrete, node-based spatial structures, such as traffic sensor networks. Their effectiveness is constrained when applied to large-scale, grid-based data, where the perceptual region is vast and dense. The sheer number of grid cells makes node-centric graph representations or direct positional encodings computationally prohibitive and inefficient, limiting their ability to perceive and model patterns over a large spatial area. Unlike existing methods, we neither limit ourselves to merely adding position encoding to traffic data without compression, nor do we need to significantly modify model structures to design specialized encoders for traffic data compression. Instead, we leverage visual encoders to transform 2D spatiotemporal sequence information based on geographic grids into contextual inputs for Vision-LLM, a strategy designed to fully unleash their inherent reasoning and generalization capabilities.

\section{2D Spatiotemporal Traffic Forecasting Problem}
To formally define the 2D spatiotemporal traffic forecasting problem, we model the urban area as a regular geographical grid of size $H \times W$, where $H$ and $W$ are the height and width of the grid, respectively. The mobile network usage across this grid at a given time step is captured as a traffic matrix. The core task is to forecast the traffic matrices for the next $K$ future time steps, given a history of observed traffic matrices over the past $S$ time steps. We denote the complete sequence of traffic data observed over a time interval of length $T$ as $\mathcal{D}=\{D_{1}, D_{2}, \dots, D_{T}\}$, where $D_t$ represents a snapshot of mobile traffic at time $t$ over the $H \times W$ geographical grid, specifically expressed as:
\begin{equation}
D_t = \begin{bmatrix}
d_t^{(1,1)} & \cdots & d_t^{(1,W)} \\
\vdots & \ddots & \vdots \\
d_t^{(H,1)} & \cdots & d_t^{(H,W)}
\end{bmatrix},
\end{equation}
where $d_{t}^{(x,y)}$ denotes the network usage metric value at geographical grid coordinates $(x,y)$, such as internet bandwidth, call frequency, or short message service (SMS) transmission rates. The spatiotemporal traffic prediction task involves forecasting mobile communication traffic across all geographical grids for the next $K$ time steps, given $S$ observed values of $D_{t}$. This can be formulated as:
\begin{equation}
\tilde{\mathbf{D}}_{t+1:t+K} = \arg\max_{\mathbf{D}_{t+1:t+K}}\, p\left(\mathbf{D}_{t+1:t+K} \mid \mathbf{D}_{t-S+1:t}\right),
\end{equation}
where $\mathbf{D}_{t+1:t+K} = \{\mathbf{D}_{t+1}, \mathbf{D}_{t+2}, \dots, \mathbf{D}_{t+K}\}$ represents the sequence of future traffic matrices across all grid cells, spanning the time interval from $t+1$ to $t+K$, i.e., the forecast target for the next $K$ time steps. Given the inherent output-length constraints of current LLMs, our approach encodes the full spatiotemporal history $\mathbf{D}_{t-S+1:t} = \{\mathbf{D}_{t-S+1}, \dots, \mathbf{D}_t\}$ while generating forecasts on a per-cell basis. To make this formulation scalable for large 2D grids, we adopt a conditional-independence approximation, inspired by the conditional factorization in Conditional Neural Processes \cite{ref_26}, and factorize the future grid distribution by conditioning each cell-level prediction on the shared global history $\mathbf{D}_{t-S+1:t}$ and its own coordinates $(x, y)$. Under this approximation, the future of each cell is primarily determined by the shared global history and the target coordinates, which leads to the following decomposition:
\begin{equation}
p(\mathbf{D}_{t+1:t+K}\!\! \mid\!\! \mathbf{D}_{t-S+1:t}) \! \approx \!\! \prod_{x=1}^{H} \! \prod_{y=1}^{W} \! p\big(\mathbf{d}^{(x,y)}_{t+1:t+K}\!\! \mid \!\!\mathbf{D}_{t-S+1:t}, \!(x,y)\big),
\end{equation}
This approximation preserves the major spatial dependencies carried by the shared global history, but it does not explicitly model the residual synchronous coupling among future cells after conditioning on that history.
Consequently, the original task can be decomposed into multiple cell-level prediction subtasks. For any coordinates $(x,y)$ in the geographical grid, where $x \in \{1,2,\dots,H\}$ and $y \in \{1,2,\dots,W\}$, the cell-level prediction solved by the LLM can be formulated as:
\begin{equation} \label{eq:prob_formulation_cell}
\tilde{\mathbf{d}}_{t+1:t+K}^{(x,y)} = 
\arg\max_{\mathbf{d}_{t+1:t+K}^{(x,y)}}\,
p\left(\mathbf{d}_{t+1:t+K}^{(x,y)} \mid \mathbf{D}_{t-S+1:t},\, (x,y)\right),
\end{equation}
Here, $\mathbf{d}_{t+1:t+K}^{(x,y)} = \{d_{t+1}^{(x,y)}, d_{t+2}^{(x,y)}, \dots, d_{t+K}^{(x,y)}\}$ represents the sequence of future traffic values for the single cell at coordinates $(x,y)$ over the next $K$ time steps. By performing predictions across all $H \times W$ geographical grids, we can reconstruct the complete spatiotemporal traffic sequence:
\begin{equation}
    \tilde{D}_{t+k} =
    \begin{bmatrix}
        \tilde{d}_{t+k}^{(1,1)} & \cdots & \tilde{d}_{t+k}^{(1,W)} \\
        \vdots & \ddots & \vdots \\
        \tilde{d}_{t+k}^{(H,1)} & \cdots & \tilde{d}_{t+k}^{(H,W)}
    \end{bmatrix},
    \quad k=1,2,\ldots,K ,
\end{equation}
where $\tilde{D}_{t+k}$ represents the predicted value of ${D}_{t+k}$.

\section{Methodology}

\begin{figure*}[!t]
\centering
\includegraphics[width=\textwidth]{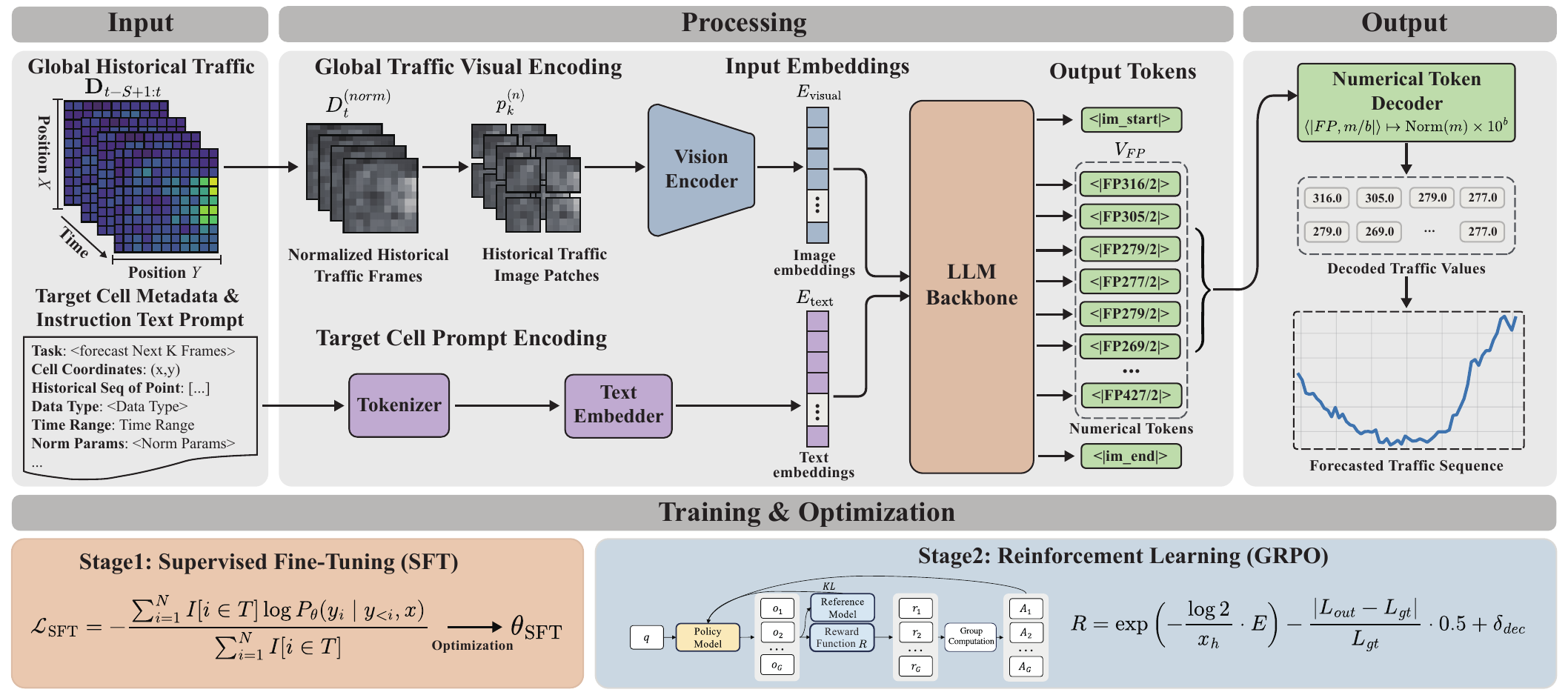}
\caption{\textbf{The ST-Vision-LLM framework.} The model takes global historical traffic frames together with target-cell metadata and textual instructions as input. The historical traffic frames are normalized, converted into image patches, and encoded by the Vision-LLM visual encoder, while the target-cell prompt is tokenized and embedded through the text branch. The resulting visual and textual embeddings are jointly processed by the LLM backbone to generate future traffic in the form of numerical tokens, which are then decoded into readable traffic values and the final forecasted traffic sequence. The lower part of the figure illustrates the two-stage optimization process consisting of supervised fine-tuning and GRPO-based reinforcement learning.}
\label{fig:framework}
\end{figure*}

\subsection{The ST-Vision-LLM Framework}
The ST-Vision-LLM framework formulates city-wide 2D spatiotemporal traffic prediction as a vision-language task. The proposed method predicts future $K$-step traffic at the cell level, conditioned on a sequence of $S$ historical frames of global traffic information $\mathbf{D}_{t-S+1:t}$. Formally, the objective for each target cell with geographic coordinates $(x,y)$ is to solve the optimization problem defined in Equation (\ref{eq:prob_formulation_cell}).

As illustrated in Figure~\ref{fig:framework}, the overall pipeline contains two input branches and two downstream optimization stages. First, the multimodal input module constructs a joint context from global traffic frames and target-cell prompt information. The historical spatiotemporal traffic sequence $\mathbf{D}_{t-S+1:t}$ is normalized, converted into image patches, and processed by the visual encoder of a foundation Vision-LLM, yielding a sequence of visual embeddings. Concurrently, the associated metadata for the target cell $(x, y)$, together with its historical scalar sequence and task instructions, are formatted into a textual prompt, tokenized, and embedded through the text branch. These visual and textual embeddings are then concatenated and consumed jointly by the LLM backbone. The complete instruction template used in training and inference is provided in Appendix~\ref{app:prompt-template}.

Second, to overcome the inherent inefficiency of LLMs in processing continuous numerical data, we introduce an efficient numerical encoding scheme. This scheme features a specialized floating-point vocabulary and an alignment fine-tuning strategy, enabling the representation of numerical values as single tokens. The LLM therefore outputs the predicted traffic sequence $\tilde{\boldsymbol{d}}_{t+1:t+K}^{(x,y)}$ as specialized numerical tokens, which are subsequently decoded into readable traffic values by a numerical token decoder. This design conserves context length while keeping the final prediction sequence directly interpretable.

Finally, we employ a two-stage training process for the optimization of predictive performance. The model first undergoes SFT to learn the fundamental spatiotemporal dynamics from the traffic data. Subsequently, the model is further refined via policy optimization, specifically using GRPO, to enhance its predictive capabilities and final performance.

\subsection{Multimodal Input Construction}
Our method embeds the historical traffic information $\mathbf{D}_{t-S+1:t}$ into the LLM context through the image encoder of a Vision-LLM, allowing the model to perceive the full spatial context of recent traffic dynamics across the entire geographical grid. Specifically, we convert the historical traffic matrices into images, segment them into patches, and feed them into the visual encoder to obtain visual embeddings. In parallel, the tokenizer and text embedding layers encode the cell-specific prompt. The autoregressive LLM then jointly consumes the visual and textual embeddings to generate numerical tokens, which are finally decoded into future traffic values.

Since traffic data exhibits long-tail distribution characteristics while image encoders typically require input data within the $[0,1]$ range, we first perform normalization on the traffic data matrix $D_t$. For this purpose, we adopt Power-Law Normalization to obtain the normalized traffic matrix $D^{(norm)}_t$. This approach first applies a power transformation with an exponent $p$ to each grid value to suppress the long tail and reduce the skewness of the data distribution \cite{ref_33}. Then, the result is divided by $t_{max}^p$ to strictly map the values into the $[0,1]$ range. Here, $p \in (0,1]$ is a hyperparameter that controls the tail compression strength. The formula is as follows:
\begin{equation}
D_t^{(norm)} = \left[\frac{(D_t^{(i,j)})^p}{t_{max}^p}\right]_{i,j},
\end{equation}
This normalization operation is performed on each input traffic data matrix to obtain the normalized global historical traffic information $\{D_{t-S+1}^{(norm)}, \ldots, D_t^{(norm)}\}$.

The traffic matrix is intrinsically a single-channel two-dimensional scalar field rather than a natural image, and it does not correspond to three different traffic variables mapped to the RGB channels. In our implementation, the same normalized traffic matrix is replicated across all three channels, so the visual encoder receives a grayscale pseudo-RGB representation instead of a semantically colored image. This design is adopted for two practical reasons only: to satisfy the three-channel input interface of the built-in visual encoder in Qwen2.5-VL \cite{ref_41}, and to avoid introducing artificial cross-channel correlations that would arise if unrelated quantities were forcibly mapped to different color channels. The motivation for using a visual encoder is not natural-image color semantics, but its ability to model patch-based two-dimensional grid structure, local neighborhood patterns, and long-range spatial dependencies \cite{ref_44}. Replicating single-channel data into three channels is also a common adaptation practice in domains such as medical imaging \cite{ref_34}. We thus construct a grayscale pseudo-RGB image format:
\begin{equation}
I_t = [D_t^{(norm)}, D_t^{(norm)}, D_t^{(norm)}],
\end{equation}
where $I_t \in \mathbb{R}^{H \times W \times 3}$ represents the three-channel image corresponding to time step $t$, with $H$ and $W$ being the spatial dimensions of the traffic matrix. For any pixel at position $(i,j)$ in the image, the RGB channel values are identical to normalized traffic values:
\begin{equation}
I_t[i,j,c] = D_t^{(norm)}[i,j], \quad c \in \{R,G,B\},
\end{equation}
where the index $c$ iterates over the Red, Green, and Blue color channels. Through this approach, all normalized traffic matrices within the historical time window are converted into an image sequence $\{I_{t-S+1}, I_{t-S+2}, \ldots, I_t\}$.

Next, we segment the image sequence into patches and process them through the Vision-LLM's image encoder. Setting the patch size of the image encoder to L×L, each image $I_t \in \mathbb{R}^{H \times W \times 3}$ can be segmented into N patches, where:
\begin{equation}
N = \lceil H/L \rceil \times \lceil W/L \rceil,
\end{equation}
For each image $I_k$ (where $k \in \{t-S+1, \ldots, t\}$) in the image sequence $\{I_{t-S+1}, I_{t-S+2}, ..., I_t\}$, we segment it into a patch sequence:
\begin{equation}
\{\mathbf{p}_k^{(1)}, \mathbf{p}_k^{(2)}, ..., \mathbf{p}_k^{(N)}\},
\end{equation}
where each patch $\mathbf{p}_k^{(n)} \in \mathbb{R}^{L \times L \times 3}$.

These patch sequences are then processed by the Vision-LLM's visual encoder. This operation, which we denote as $\text{ImageEncoder}(\cdot)$, transforms the patch sequence of a single image $I_k$ into a corresponding sequence of visual embeddings $E_k$:
\begin{equation}
E_k = \text{ImageEncoder}(\{\mathbf{p}_k^{(1)}, \mathbf{p}_k^{(2)}, ..., \mathbf{p}_k^{(N)}\}),
\end{equation}
To construct the complete visual context for the entire historical window, these embedding sequences are concatenated in temporal order. This results in $\mathbf{E}_{\text{visual}}$, a single flattened sequence of embeddings that represents the global spatiotemporal traffic history:
\begin{equation}
\mathbf{E}_{\text{visual}} = [E_{t-S+1}, E_{t-S+2}, ..., E_t],
\end{equation}
Finally, these visual embeddings are concatenated with the text prompt embeddings to form the complete input context for the LLM:
\begin{equation}
\mathbf{Context} = [\mathbf{E}_{\text{visual}}, \mathbf{E}_{\text{text}}],
\end{equation}
where $\mathbf{E}_{\text{text}}$ represents the embeddings of the text instructions, and $\mathbf{E}_{\text{visual}} \in \mathbb{R}^{(S \times N) \times d}$ contains the complete 2D spatiotemporal traffic information from the historical window, with $d$ being the embedding dimension.

\subsection{Direct Numerical Encoding Method}

\subsubsection{Numerical Token Design}
A significant challenge in applying LLMs to numerical tasks is their inherent inefficiency in processing floating-point numbers, which are typically tokenized into multiple, less meaningful character-level tokens. To overcome this limitation and enhance both training and inference efficiency by compressing sequence length, we propose a direct numerical encoding method for the efficient encoding of floating-point numbers into single tokens, thereby representing real numbers in a discrete vocabulary space. This method is inspired by Charton (2022) \cite{ref_27}, and its primary advantage lies in its ability to represent floating-point numbers of varying precision as a single token, thus significantly compressing the sequence length of numerical data in the model's input.

In particular, we construct a dedicated numerical vocabulary, $V_{FP}$, composed of a series of special numerical tokens of the form $\langle|\text{FP } m/b|\rangle$. Here, ``FP" is an abbreviation for ``Floating-Point," serving to identify the string as a special token that represents a floating-point value. In this format, $m$ represents the integer mantissa, and $b$ represents the base-10 exponent. The mapping from these numerical tokens to their corresponding floating-point values is defined as:
\begin{equation}
\langle|\text{FP } m/b|\rangle \mapsto \text{Norm}(m) \times 10^b,
\end{equation}
Here, $\text{Norm}(m)$ is a normalization function that converts the integer $m$ into its scientific notation form by placing the decimal point after the first significant digit. This function decouples the magnitude of the final floating-point number from the magnitude of $m$, ensuring that the magnitude of the final floating-point number is determined solely by the exponent $b$. This operation is formally defined as:
\begin{equation}
\text{Norm}(m) = \frac{m}{10^{\lfloor \log_{10}|m| \rfloor}}, \quad m \neq 0,
\end{equation}
The zero value is handled separately by the dedicated token $\langle|\text{FP0/0}|\rangle$. We set the range for $m$ to $\{-999, \dots, -1\} \cup \{1, \dots, 999\}$ and the range for the exponent $b$ to $\{-4, \dots, 5\}$, yielding a three-digit mantissa precision while still covering the numerical dynamic range common in our task based on the numerical distribution of traffic data in the Telecom Italia Big Data Challenge \cite{ref_38} dataset. By iterating through all possible $(m,b)$ combinations, we construct a numerical vocabulary that covers a wide dynamic range.

We define a single token representing a floating-point number encoded via this method as a \textbf{numerical token}. Compared to representing numbers as direct string sequences, this single-token encoding method significantly reduces sequence length, thereby enhancing the model's training and inference efficiency.

\subsubsection{Numerical Alignment Fine-tuning Method}
Having designed a specialized vocabulary for numerical values, the next critical step is to imbue the pre-trained LLM with the ability to comprehend and generate these new tokens. Unlike Charton (2022) \cite{ref_27}, who trained LLMs from scratch using pure numerical tokens for linear algebra tasks, our research focuses on fine-tuning pre-trained LLMs to understand and generate numerical tokens. A numerical alignment fine-tuning method is proposed to realize this goal, enabling conventional pre-trained LLMs to acquire the capability of processing numerical tokens and executing downstream tasks.

The numerical fine-tuning method comprises two stages: semantic alignment fine-tuning and basic arithmetic alignment fine-tuning. The first stage aligns $V_{FP}$ with text-represented floating-point numbers in the embedding space, and the second stage trains the LLM to perform basic operations with numerical tokens using mixed-format linear algebra data.

\textbf{Stage 1: Semantic Alignment Fine-tuning}

Having established the vocabulary, the immediate challenge is to train the newly initialized parameters within the word embedding layer, enabling the model to comprehend these new tokens. The primary goal of this stage is to bridge the semantic gap between our newly introduced numerical tokens and the model's pre-existing understanding of numbers represented as text. To achieve this, we first extend the LLM’s input and output embedding matrices to include $V_{\mathrm{FP}}$.  Then align the newly initialized numeric embeddings with their textual numeral counterparts so that, for instance, $\langle|\text{FP114/0}|\rangle$ and the string ``1.14'' occupy nearby positions in the semantic space. This process ensures that the model interprets the new numerical token not as an arbitrary symbol, but as a meaningful representation of a quantity, semantically grounding it within the model's existing understanding of numbers. To this end, we freeze the backbone and fine-tune only the input/output embeddings on transcription tasks, establishing a bidirectional mapping between textual numerals and numeric tokens. Example prompts:
\begin{itemize}
    \item ``Convert the following string-represented numerical value to numerical tokens: `-82.100000' $\rightarrow$ $\langle|\text{FP-821/1}|\rangle$''
    \item ``Transcribe the following numerical token to string numerical value: $\langle|\text{FP114/2}|\rangle$ $\rightarrow$ `114.000000' ''
\end{itemize}
Through training in this stage, numerical tokens achieve basic alignment with text-represented floating-point numbers in semantic space, acquiring reasonable numerical representation capabilities.

\textbf{Stage 2: Basic Arithmetic Alignment Fine-tuning}

Stage 2 equips the LLM with basic arithmetic abilities needed to handle numerical tokens in traffic forecasting. We therefore continue fine-tuning on numerical computation tasks by unfreezing the backbone, token embedding layer, and output layer. Since the backbone is already pre-trained and the new embeddings have been preliminarily aligned, we adopt Low-Rank Adaptation (LoRA) \cite{ref_28} for parameter-efficient fine-tuning of these components.

This fine-tuning is applied to a preexisting LLM.  Given that pre-trained models already possess foundational mathematical abilities \cite{ref_29}, our training does not involve complex mathematical tasks, but rather concentrates on three fundamental linear algebra operations that can effectively inject numerical capabilities through a few templated examples \cite{ref_30} while respectively covering key mechanisms in the Transformer architecture, such as approximate linear estimation \cite{ref_31} and native element-wise interactions \cite{ref_32}: vector addition, vector subtraction, and the Hadamard product.
For vectors $\mathbf{a}=[a_1,\ldots,a_n]$ and $\mathbf{b}=[b_1,\ldots,b_n]$:
\begin{itemize}
    \item Addition: $\mathbf{a}+\mathbf{b} = [a_1 + b_1, \ldots, a_n + b_n]$.
    \item Subtraction: $\mathbf{a}-\mathbf{b} = [a_1 - b_1, \ldots, a_n - b_n]$.
    \item Hadamard Product: $\mathbf{a} \odot \mathbf{b} = [a_1 b_1, \ldots, a_n b_n]$.
\end{itemize}
Each task includes a combination of three input-output formats: numerical tokens to numerical strings, numerical strings to numerical tokens, and numerical tokens to numerical tokens. The numerical-token-to-string format is used only for semantic alignment and intermediate arithmetic supervision, while the final traffic forecasting task still outputs numerical tokens. This design ensures the model can flexibly convert between different numerical representations and perform the corresponding mathematical operations. During task construction, we ensure that the numerical tokens involved in the tasks completely cover the entire numerical vocabulary $V_{FP}$.
Through the two-phase fine-tuning tasks, the model acquires the ability to understand, manipulate, and generate sequences of numerical tokens, providing an efficient numerical encoding method for the subsequent traffic forecasting task.

\subsection{SFT and Reinforcement Learning Optimization}
After numerical alignment fine-tuning and multimodal input construction, the model has acquired the capability to process numerical tokens and understand global 2D spatiotemporal traffic information. To maximize predictive accuracy and ensure the generation of structurally correct outputs in traffic prediction tasks, our method employs a two-stage training strategy. The first stage uses SFT to establish a foundational predictive capability. In this phase, the model learns to imitate the ground-truth data, aligning its generative process with the task of mapping global historical traffic and cell-specific prompts to a corresponding future traffic sequence. The second stage then employs the GRPO method \cite{ref_16} to optimize this learned prediction strategy, further improving prediction accuracy.

\subsubsection{SFT Fine-tuning Stage}
In the SFT stage, we align the model's generative process with the prediction task by training it on input-output pairs. For each training sample, we construct a single sequence by concatenating the complete input context with the corresponding ground-truth output sequence. The input context includes the image patches representing global historical traffic, the target cell's geographical coordinates $(x,y)$ as previously defined, its historical traffic data, and other metadata.

The model is then trained using a causal language modeling objective with \textit{teacher-forcing}. This involves maximizing the likelihood of the ground-truth tokens in the output sequence, conditioned on the preceding ground-truth tokens and the input context. Let $\mathcal{T}$ be the set of indices corresponding to the tokens of the target output sequence. The loss is calculated using the cross-entropy objective, but only over the tokens in $\mathcal{T}$. This technique, often used in instruction-tuning \cite{ref_37}, is formalized as
\begin{equation}
\mathcal{L}_{\text{SFT}} = -\frac{\sum_{i=1}^{N}\mathbb{I}[i\in \mathcal{T}]\,\log P_\theta(y_i\mid y_{<i},x)}{\sum_{i=1}^{N}\mathbb{I}[i\in \mathcal{T}]},
\end{equation}
where $N$ is the total sequence length, $x$ is the input context, $y_i$ is the token at position $i$, and $\theta$ represents the model parameters. The term $P_\theta$ denotes the conditional probability distribution over the vocabulary predicted by the model. The indicator function $\mathbb{I}[i \in \mathcal{T}]$ implements loss masking. It equals 1 if the token at position $i$ is part of the desired output, and 0 otherwise. This ensures that the loss is computed exclusively on the model's predictions for the target sequence, ignoring the input context portion.

\subsubsection{GRPO Fine-tuning Stage}
Upon completing the SFT stage, the model has acquired basic traffic prediction capabilities, but its prediction strategy may not yet be optimal. To further enhance prediction accuracy, we employ a reinforcement learning optimization stage. Unlike traditional supervised learning that relies solely on historical labels, reinforcement learning allows for the direct optimization of prediction metrics. We specifically choose GRPO \cite{ref_16}, a memory-efficient variant of Proximal Policy Optimization (PPO)\cite{ref_17}. Standard PPO is an actor-critic method that requires training a separate, computationally expensive critic network to estimate the value of states, which serves as a baseline for calculating advantages. GRPO innovatively foregoes this critic network, making it a more practical choice for fine-tuning large models.

GRPO estimates advantages through group-wise comparisons. For each input prompt, ST-Vision-LLM generates a group of $G$ candidate output sequences, each of which is scored by the reward function. Instead of using a critic network, GRPO computes the baseline directly from the group performance, typically as the average reward, and defines the advantage of each sequence relative to this group average.

Finally, the policy network is updated to increase the likelihood of generating sequences that achieve higher-than-average rewards. The GRPO objective function optimizes the policy by maximizing this group-relative advantage. To maintain training stability and prevent the policy from deviating too drastically from the well-behaved SFT model, the update is regularized by a KL-divergence penalty against a frozen reference model. The reference model is typically the initial SFT-tuned model. This entire process guides the model to learn a superior prediction strategy by directly optimizing for higher reward scores.

In our traffic prediction task, we define the reward function $R$ as:
\begin{equation} \label{eq:R_score}
\begin{split}
R ={}& \exp\left(-\frac{\log 2}{x_{h}}\cdot \mathcal{E}\right) 
     - \frac{\lvert L_{\text{out}} - L_{\text{gt}} \rvert}{L_{\text{gt}}}\cdot 0.5 + \delta_{\text{dec}},
\end{split}
\end{equation}
where $\delta_{\text{dec}}$ is a penalty term for decoding failures, given by:
\begin{equation} \label{eq:delta_decode}
\delta_{\text{dec}} =
\begin{cases}
0, & \text{successful decoding} \\
-0.5, & \text{decoding failure}
\end{cases}
\end{equation}
The term $\mathcal{E}$ represents the Normalized Root Mean Square Error (NRMSE), which we define as:
\begin{equation} \label{eq:nrmse}
\mathcal{E} = \frac{1}{d} \sqrt{\frac{1}{N} \sum_{k=1}^{N} (\hat{d}_k - d_k)^2}.
\end{equation}
In these formulas, $x_h$ is the half-score rate, a hyperparameter ensuring the reward is 0.5 when the error $\mathcal{E}$ equals $x_h$. The term $\exp(-\frac{\log 2}{x_h} \cdot \mathcal{E})$ is the accuracy reward component, valued in $[0,1]$. $L_{gt}$ is the length of the ground-truth sequence, and $L_{out}$ is the length of the model's output sequence. The term $- \frac{| L_{\text{out}} - L_{\text{gt}} |}{L_{\text{gt}}} \cdot 0.5$ is a length mismatch penalty. In the NRMSE formula, $N$ is the number of prediction steps, $\hat{d}_k$ and $d_k$ are the predicted and ground-truth values at step $k$, respectively, and $d$ is a normalization constant. Training is terminated when the NRMSE on the validation set ceases to decrease, which yields the final trained model.

\section{Experiments}
In this section, we conduct a series of experiments to comprehensively evaluate the effectiveness of ST-Vision-LLM on spatio-temporal sequence forecasting tasks. Specifically, we first validate its benchmark performance on long-term and short-term prediction, as well as cross-domain prediction tasks. Subsequently, we further investigate the model's performance in data-scarce and knowledge-transfer scenarios, such as few-shot learning, cross-domain few-shot fine-tuning, and zero-shot prediction, to fully demonstrate the few-shot learning and generalization capabilities of ST-Vision-LLM.

\subsection*{Mobile Traffic Dataset}
This study utilizes a public, real-world mobile traffic dataset originally released for the Telecom Italia Big Data Challenge \cite{ref_38}. The dataset contains measurements of total cellular traffic activity in the city of Milan and the Trentino province from November 1, 2013, to January 1, 2014 (a total of two months), at a 10-minute observation interval. These two geographical areas differ significantly in nature and population density, thus exhibiting distinct traffic patterns. Specifically, the Milan area is divided into a $100 \times 100$ grid of cells, where each cell covers an area of approximately 55,000 square meters (i.e., $235\text{m} \times 235\text{m}$). In contrast, the Trentino area consists of a $117 \times 98$ grid of cells, with each cell covering an area of 1,000,000 square meters.

\subsection*{Baseline Models}
We compare ST-Vision-LLM against the following 12 baseline models, which can be categorized into four groups: (1) statistical models: ARIMA \cite{ref_18}; (2) models based on recurrent or convolutional neural networks: STN \cite{ref_6}, ST-ResNet \cite{ref_20}, ResLSTM \cite{ref_19}, ACFM \cite{ref_24}; (3) models based on graph neural networks: STGCN \cite{ref_21}, MCSTGCN \cite{ref_23}, GWNET \cite{ref_22}; (4) models based on LLMs: ST-LLM \cite{ref_12}, GCNGPT \cite{ref_12}, GATGPT \cite{ref_25}, Time-LLM \cite{ref_10}. The details of these baseline models are as follows:
\begin{itemize}
    \item ST-LLM \cite{ref_12} defines the time steps at each location as tokens and designs a spatio-temporal embedding to learn the spatial positions and global temporal patterns of these tokens.
    \item GCNGPT \cite{ref_12} combines a graph convolutional network with a frozen pre-trained Transformer.
    \item GATGPT \cite{ref_25} fuses a graph attention mechanism with an LLM to enhance the understanding of spatial relationships.
    \item Time-LLM \cite{ref_10} reprograms the input time series via textual prototypes to align it with a frozen LLM.
    \item ACFM \cite{ref_24} employs an attention mechanism and consists of two progressive ConvLSTM units.
    \item GWNET \cite{ref_22} introduces an adaptive adjacency matrix and combines it with stacked dilated 1D convolutions.
    \item ResLSTM \cite{ref_19} combines a residual network, a graph convolutional network, and a long short-term memory network.
    \item ST-ResNet \cite{ref_20} leverages a residual neural network framework to model temporal closeness, period, and trend.
    \item STGCN \cite{ref_21} integrates graph convolutional layers with convolutional sequence learning layers to handle spatio-temporal data.
    \item MCSTGCN \cite{ref_23} adopts a multi-component approach to separately capture spatio-temporal correlations within different time periods.
    \item STN \cite{ref_6} fuses ConvLSTM and 3D-ConvNet architectures to extract spatio-temporal features.
    \item ARIMA \cite{ref_18} is an autoregressive integrated moving average model with a Kalman filter.
\end{itemize}

\subsection*{Data Preprocessing and Experimental Setup}
Due to data storage errors or improper transmission, some cellular cells have missing data for certain time periods, resulting in discontinuous traffic data. Therefore, data imputation is necessary before performing traffic forecasting. We employ \textbf{linear interpolation} to fill these gaps, which preserves temporal continuity without introducing strong distributional assumptions.
For any grid cell $(x,y)$, if its time series contains a sequence of consecutive missing values, we use linear interpolation for imputation. For example, for a sequence of missing values from time $t_{start}$ to $t_{end}$, where $d_{t_{start}}^{(x,y)}$ and $d_{t_{end}}^{(x,y)}$ are known non-missing traffic values, the traffic value $\tilde{d}_t^{(x,y)}$ at any time $t$ between $t_{start}$ and $t_{end}$ (i.e., $t_{start} < t < t_{end}$) is calculated as:
\begin{equation}
\tilde{d}_t^{(x,y)} = d_{t_{start}}^{(x,y)} + \frac{t - t_{start}}{t_{end} - t_{start}} \left(d_{t_{end}}^{(x,y)} - d_{t_{start}}^{(x,y)}\right),
\end{equation}
After data imputation, the model is trained using the complete imputed traffic data, which serves as both model input and training labels. However, during the model validation phase, treating imputed data as ground truth would introduce bias, compromising a fair evaluation. Therefore, although the model input still contains the imputed traffic data, the error computation for the output is performed only on the original observed data points, excluding the interpolated parts.

We train all deep learning models on data from November 1, 2013, to December 18, 2013 (48 days), validate them on the next 7 days of data, and then evaluate their performance on data from December 26, 2013, to January 1, 2014. We use Mean Absolute Error (MAE), Root Mean Squared Error (RMSE), and NRMSE to quantify the accuracy of our proposed ST-Vision-LLM and existing methods. They are defined as:
\begin{gather}
E_{\text{mae}} = \frac{1}{n} \sum_{i=1}^{n} |y_i - \hat{y}_i| \label{eq:mae}, \\
E_{\text{rmse}} = \sqrt{ \frac{1}{n} \sum_{i=1}^{n} (y_i - \hat{y}_i)^2 } \label{eq:rmse}, \\
E_{\text{nrmse}} = \frac{E_{\text{rmse}}}{\bar{y}} = \frac{ \sqrt{ \frac{1}{n} \sum_{i=1}^{n} (y_i - \hat{y}_i)^2 } }{ \frac{1}{n} \sum_{i=1}^{n} y_i } \label{eq:nrmse_mean},
\end{gather}
Where $\hat{y}_i$ is the predicted value, $y_{i}$ is its corresponding ground truth value, and $n$ is the total number of observed values in the sample. The metric $E_{\text{nrmse}}$ is commonly used to evaluate model performance across datasets with different scales or units, as it normalizes the RMSE by the mean of the ground truth values, thereby eliminating the effect of scale. A lower metric value indicates a more accurate prediction.

Given computational cost constraints, in all datasets, we conduct training and validation only on the central region $(x, y) \in [45, 55) \times [45, 55)$. This means the model is tasked with predicting the output for this central sub-region only, although it can still perceive information from the entire input region. The experiments were conducted on an NVIDIA 5090 GPU with 32GB of memory.

To comprehensively compare the performance of ST-Vision-LLM with existing methods, we use five channels from the original dataset: Internet, sms\_in, sms\_out, call\_in, and call\_out. We select sms\_in for the SMS category and call\_in for the Call category. This yields six distinct telecommunication spatio-temporal traffic subsets, named Milan-Internet, Milan-SMS, Milan-Call, Trentino-Internet, Trentino-SMS, and Trentino-Call \cite{ref_38}. All subsequent tests are performed on these six subsets. In the benchmark and few-shot tests, we train and test the models on the SMS and Internet subsets. In the transfer learning and zero-shot tests, we first train the models on the Call subset and then perform transfer training and validation on the Internet and SMS subsets.

For methods using LLMs, Time-LLM \cite{ref_10} uses Qwen2.5-7B \cite{ref_40} as its backbone model, while our ST-Vision-LLM method uses Qwen2.5-VL-7B-Instruct \cite{ref_41} as the base model for fine-tuning. In ST-Vision-LLM, the historical traffic frames are first processed by the built-in visual encoder of the Vision-LLM to obtain visual embeddings, and these visual embeddings are then fed, together with the textual instruction, into the language model for autoregressive prediction. Its image encoder is the native visual tower of that base model rather than an external visual front-end. As for the other comparative methods, we used the default configurations reported in their respective papers.

\begin{table*}[!t]
\centering
\caption{Results of long-term and short-term baseline prediction. Trained and evaluated on Milan-Internet, with the input sequence length set to $S=12$ and prediction horizons set to $K \in \{1, 10, 30, 60\}$. The best result is marked in \textbf{bold} and the second best is \underline{underlined}. }
\label{tab:milan_internet_comparison_updated}
\resizebox{\textwidth}{!}{%
\begin{tabular}{l|rrr|rrr|rrr|rrr}
    \hline
    \multicolumn{1}{c|}{\textbf{Horizon}} & \multicolumn{3}{c|}{\textbf{1-step}} & \multicolumn{3}{c|}{\textbf{10-step}} & \multicolumn{3}{c|}{\textbf{30-step}} & \multicolumn{3}{c}{\textbf{60-step}} \\
    \cline{2-4} \cline{5-7} \cline{8-10} \cline{11-13}
    \multicolumn{1}{c|}{\textbf{Metric}} & \multicolumn{1}{c}{MAE} & \multicolumn{1}{c}{RMSE} & \multicolumn{1}{c|}{NRMSE} & \multicolumn{1}{c}{MAE} & \multicolumn{1}{c}{RMSE} & \multicolumn{1}{c|}{NRMSE} & \multicolumn{1}{c}{MAE} & \multicolumn{1}{c}{RMSE} & \multicolumn{1}{c|}{NRMSE} & \multicolumn{1}{c}{MAE} & \multicolumn{1}{c}{RMSE} & \multicolumn{1}{c}{NRMSE} \\
    \hline
    ST-Vision-LLM & \underline{21.6525} & 31.8887 & 0.1018 & \textbf{27.4897} & \textbf{41.8479} & \textbf{0.1336} & \textbf{47.3410} & \textbf{70.9002} & \textbf{0.2258} & \textbf{75.8858} & \textbf{103.8315} & \textbf{0.3297} \\
    ST-LLM        & 30.2360 & 41.5694 & 0.1325 & 51.5147 & 72.4567 & 0.2307 & 90.0111 & 149.3411 & 0.4747 & 109.3175 & 191.8308 & 0.6092 \\
    GCNGPT        & 29.5151 & 40.3999 & 0.1287 & 43.3770 & 58.4880 & 0.1862 & 77.5356 & 120.7583 & 0.3838 & 109.4818 & 195.0954 & 0.6195 \\
    GATGPT        & 30.3270 & 41.5221 & 0.1323 & 46.5658 & 65.2929 & 0.2079 & 89.1961 & 144.1842 & 0.4583 & 122.4923 & 208.4570 & 0.6619 \\
    Time-LLM      & 33.0400 & 45.3200 & 0.1447 & 56.7800 & 78.6500 & 0.2504 & 94.3400 & 151.9100 & 0.4828 & 132.4700 & 231.1900 & 0.7341 \\
    ACFM          & 28.1887 & 36.8261 & 0.1176 & 48.1180 & 76.4256 & 0.2433 & 79.0057 & 146.0898 & 0.4643 & 109.5715 & 183.7074 & 0.5833 \\
    GWNET         & \textbf{14.9597} & \textbf{23.2448} & \textbf{0.0746} & \underline{27.8492} & \underline{42.5134} & \underline{0.1353} & 76.2321 & 140.6652 & 0.4471 & 103.8087 & 180.1602 & 0.5721 \\
    ResLSTM       & 34.8413 & 49.5176 & 0.1581 & 41.7371 & 59.8353 & 0.1905 & \underline{63.9126} & \underline{85.6559} & \underline{0.2723} & 110.8396 & 147.7705 & 0.4696 \\
    ST-ResNet     & 34.2675 & 46.4401 & 0.1480 & 49.3016 & 67.6645 & 0.2154 & 97.5731 & 165.3917 & 0.5257 & 139.4607 & 196.9013 & 0.6253 \\
    STGCN         & 21.6623 & \underline{28.5073} & \underline{0.0910} & 68.8386 & 91.8513 & 0.2924 & 83.3191 & 113.1243 & 0.3596 & 112.0226 & 153.3440 & 0.4870 \\
    MCSTGCN       & 56.4753 & 87.0866 & 0.2782 & 93.1341 & 137.9330 & 0.4391 & 87.9280 & 129.3152 & 0.4110 & \underline{98.7467} & \underline{144.8693} & \underline{0.4601} \\
    STN           & 39.5036 & 63.8600 & 0.2034 & 55.0212 & 91.1873 & 0.2903 & 80.7531 & 144.1812 & 0.4583 & 99.6591 & 177.8506 & 0.5648 \\
    ARIMA         & 40.2710 & 66.8839 & 0.2134 & 68.3513 & 113.3009 & 0.3607 & 125.4663 & 244.0582 & 0.7122 & 156.6109 & 291.2429 & 0.9249 \\
    \hline
\end{tabular}%
}
\end{table*}

\begin{table*}[!t]
\centering
\caption{Experimental results for cross-domain prediction baselines. The input sequence length is set to $S=12$, and the prediction horizon is set to $K=36$. The best result is marked in \textbf{bold} and the second best is \underline{underlined}.}
\label{tab:cross_domain_no_gwnet_updated}
\resizebox{\textwidth}{!}{%
\begin{tabular}{l|rrr|rrr|rrr|rrr}
    \hline
    \multicolumn{1}{c|}{\textbf{Dataset}} & \multicolumn{3}{c|}{\textbf{Trentino-Internet}} & \multicolumn{3}{c|}{\textbf{Trentino-SMS}} & \multicolumn{3}{c|}{\textbf{Milan-Internet}} & \multicolumn{3}{c}{\textbf{Milan-SMS}} \\
    \cline{2-4} \cline{5-7} \cline{8-10} \cline{11-13}
    \multicolumn{1}{c|}{\textbf{Metric}} & \multicolumn{1}{c}{MAE} & \multicolumn{1}{c}{RMSE} & \multicolumn{1}{c|}{NRMSE} & \multicolumn{1}{c}{MAE} & \multicolumn{1}{c}{RMSE} & \multicolumn{1}{c|}{NRMSE} & \multicolumn{1}{c}{MAE} & \multicolumn{1}{c}{RMSE} & \multicolumn{1}{c|}{NRMSE} & \multicolumn{1}{c}{MAE} & \multicolumn{1}{c}{RMSE} & \multicolumn{1}{c}{NRMSE} \\
    \hline
    ST-Vision-LLM & \underline{14.5816} & \textbf{30.4686} & \textbf{0.4393} & 2.9704 & \underline{6.8868} & \underline{0.8672} & \textbf{47.1007} & \textbf{66.9314} & \textbf{0.2126} & \textbf{7.2017} & \textbf{17.6204} & \textbf{0.7213} \\
    ST-LLM        & 15.8334 & 35.1622 & 0.5069 & \underline{2.8258} & 6.9871 & 0.8804 & 91.0541 & 154.3968 & 0.4904 & 8.9303 & 18.9329 & 0.7750 \\
    GCNGPT        & 17.0779 & 39.0128 & 0.5625 & 3.0729 & 7.4832 & 0.9428 & 83.8998 & 146.0534 & 0.4639 & 9.5920 & 18.8464 & 0.7715 \\
    GATGPT        & \textbf{14.5254} & 31.7930 & 0.4583 & \textbf{2.7562} & \textbf{6.6744} & \textbf{0.8410} & 103.7383 & 167.0780 & 0.5307 & \underline{8.9146} & \underline{18.5483} & \underline{0.7593} \\
    Time-LLM      & 30.2154 & 66.4529 & 0.9582 & 5.5212 & 11.2638 & 1.4193 & 93.3401 & 175.6411 & 0.5579 & 16.8151 & 29.2006 & 1.1953 \\
    ACFM          & 16.6406 & 39.3074 & 0.5668 & 6.6483 & 13.4389 & 1.6934 & 113.4153 & 184.9829 & 0.5876 & 24.3491 & 37.5989 & 1.5391 \\
    MCSTGCN       & 14.9153 & \underline{31.2775} & \underline{0.4510} & 6.2741 & 11.7457 & 1.4803 & \underline{72.8613} & \underline{115.3424} & \underline{0.3664} & 15.4320 & 26.2068 & 1.0728 \\
    STN           & 15.0197 & 32.1622 & 0.4638 & 3.9015 & 8.7322 & 1.1003 & 83.7100 & 145.7200 & 0.4629 & 16.2560 & 27.3900 & 1.1212 \\
    ARIMA         & 45.2614 & 99.5753 & 1.4358 & 5.0521 & 10.5107 & 1.3244 & 125.4663 & 228.6892 & 0.7264 & 15.7524 & 26.3374 & 1.0781 \\
    \hline
\end{tabular}%
}
\end{table*}

\subsection{Long-term and Short-term Prediction Benchmark}
\subsubsection*{Setup}
We conduct evaluations on the Milan-Internet dataset, using the entire training set for training. The input sequence length $S$ is set to 12 frames, with each frame spaced 10 minutes apart, corresponding to a 2-hour input window. We evaluate the model's performance under different prediction horizons $K$, where $K \in \{1, 10, 30, 60\}$ steps. These prediction steps correspond to 10 minutes, 1 hour 40 minutes, 5 hours, and 10 hours, respectively. The evaluation metrics are MAE, RMSE, and NRMSE.

\subsubsection*{Results}
Table~\ref{tab:milan_internet_comparison_updated} presents a performance comparison of ST-Vision-LLM against 12 baseline models across different prediction horizons. We can draw the following observations. (1) ST-Vision-LLM consistently outperforms all baseline models in multi-step and long-term prediction tasks. (2) Compared to other LLM-based baselines, ST-Vision-LLM exhibits a significant advantage. For instance, in a 60-step prediction, ST-Vision-LLM reduces the NRMSE by approximately 45\% compared to ST-LLM, demonstrating its superior capability in leveraging vision-language models for spatio-temporal forecasting. (3) When compared with state-of-the-art GNN-based models, we observe that GWNET demonstrates the strongest performance in single-step prediction. However, as the prediction horizon increases, the advantage of ST-Vision-LLM becomes apparent. In the 10-step prediction, ST-Vision-LLM's performance is already slightly superior to GWNET, and its performance advantage further widens over longer prediction horizons, namely 30 and 60 steps. (4) Compared to traditional models based on recurrent or convolutional neural networks and statistical models, ST-Vision-LLM shows an overwhelming advantage across all prediction horizons. (5) A noteworthy observation is that the performance of many advanced baseline models degrades significantly as the prediction horizon increases, whereas ST-Vision-LLM's error grows more gracefully, indicating its stronger robustness in capturing and extrapolating long-term spatio-temporal dependencies. Overall, the experimental results indicate that while ST-Vision-LLM shows competitive performance in single-step prediction, its true strength lies in multi-step and long-term forecasting, making it a powerful solution for spatio-temporal tasks that require reliable long-range predictions.

\subsection{Cross-domain Prediction Benchmark}
\subsubsection*{Setup}
To test the model's long-term prediction accuracy across different domains, we conduct evaluations on four subsets: Trentino-Internet, Trentino-SMS, Milan-Internet, and Milan-SMS. The output sequence length $K$ is set to 36 steps, corresponding to a 6-hour prediction window. Other experimental settings are consistent with those in the long-term and short-term prediction benchmark.

\subsubsection*{Results}
Table~\ref{tab:cross_domain_no_gwnet_updated} shows the cross-domain prediction results on four different subsets. We can draw the following observations. (1) ST-Vision-LLM achieves the best performance on three of the subsets and consistently demonstrates superior prediction accuracy across all four. (2) Compared to other LLM-based baselines, ST-Vision-LLM shows consistent superiority. Interestingly, GATGPT exhibits strong competitiveness on the Trentino datasets, but its performance declines on the Milan datasets, suggesting that ST-Vision-LLM has stronger generalization capability across different data domains. (3) When compared with advanced GNN-based models, ST-Vision-LLM also shows a significant advantage. For example, on the Milan-Internet dataset, ST-Vision-LLM reduces the RMSE by over 42\% compared to the second-best model, MCSTGCN. (4) Traditional statistical models and models based on recurrent or convolutional neural networks, such as ARIMA and STN, lag significantly in performance across all test scenarios, highlighting the advantages of modern approaches in handling complex spatio-temporal data. In summary, these cross-domain prediction results provide strong evidence of ST-Vision-LLM's robustness and adaptability across different geographical regions and data types, underscoring its potential as a general-purpose spatio-temporal forecasting framework.

\subsection{Few-shot Prediction}

\begin{table}[h!]
\centering
\caption{Experimental results for 10\% few-shot prediction in terms of NRMSE. The input sequence length is set to $S=12$, and the prediction horizon is set to $K=36$. Trained using 10\% of the dataset. The best result is marked in \textbf{bold} and the second best is \underline{underlined}.}
\label{tab:few_shot_10_transposed_narrow2_clean}
\begin{tabular}{lcccc}
\toprule
Method & \begin{tabular}[c]{@{}c@{}}Trentino- \\ Internet\end{tabular} & \begin{tabular}[c]{@{}c@{}}Trentino- \\ SMS\end{tabular} & \begin{tabular}[c]{@{}c@{}}Milan- \\ Internet\end{tabular} & \begin{tabular}[c]{@{}c@{}}Milan- \\ SMS\end{tabular} \\
\midrule
ST-Vision-LLM & \underline{0.6357} & \textbf{0.9882} & \textbf{0.2537} & \textbf{0.7324} \\
ST-LLM & 0.6979 & 1.3362 & 0.4555 & 0.8901 \\
GCNGPT & 0.7363 & 1.4068 & \underline{0.4410} & \underline{0.8575} \\
GATGPT & \textbf{0.5249} & \underline{1.1700} & 0.5154 & 0.8679 \\
Time-LLM & 0.9616 & 1.3615 & 0.5639 & 0.9315 \\
GWNET & 0.6495 & 1.4182 & 0.5881 & 1.1417 \\
ResLSTM & 0.8635 & 1.5129 & 0.9958 & 0.9412 \\
STN & 1.1736 & 1.3032 & 0.6696 & 0.9223 \\
\bottomrule
\end{tabular}
\end{table}

\begin{table}[h!]
\centering
\caption{Experimental results for 5\% few-shot prediction in terms of NRMSE. The input sequence length is set to $S=12$, and the prediction horizon is set to $K=36$. Trained using 5\% of the dataset. The best result is marked in \textbf{bold} and the second best is \underline{underlined}.}
\label{tab:few_shot_5_transposed}
\begin{tabular}{lcccc}
\toprule
Method & \begin{tabular}[c]{@{}c@{}}Trentino- \\ Internet\end{tabular} & \begin{tabular}[c]{@{}c@{}}Trentino- \\ SMS\end{tabular} & \begin{tabular}[c]{@{}c@{}}Milan- \\ Internet\end{tabular} & \begin{tabular}[c]{@{}c@{}}Milan- \\ SMS\end{tabular} \\
\midrule
ST-Vision-LLM & 0.7451 & \textbf{1.0505} & \textbf{0.3887} & \textbf{0.6801} \\
ST-LLM & 0.7678 & 1.5191 & 0.5100 & \underline{0.8892} \\
GCNGPT & 0.8004 & 1.4890 & \underline{0.4379} & 0.9936 \\
GATGPT & \textbf{0.6354} & \underline{1.4071} & 0.5209 & 0.9302 \\
Time-LLM & 0.9668 & 1.4622 & 0.5403 & 1.2338 \\
GWNET & \underline{0.6490} & 1.5290 & 0.7070 & 1.0813 \\
ResLSTM & 0.9310 & 1.6536 & 0.9524 & 1.0343 \\
STN & 1.6943 & 1.4863 & 1.4616 & 1.1412 \\
\bottomrule
\end{tabular}
\end{table}

\subsubsection*{Setup}
LLMs, benefiting from their extensive pre-training data, often exhibit excellent few-shot learning capabilities \cite{ref_29}. This ability is particularly advantageous in data-scarce scenarios. In this section, we follow the configuration of the cross-domain prediction benchmark. Building on this, we evaluate the few-shot learning capability of each method by limiting the amount of training data (i.e., using only the first 10\% and 5\% of the training data).

\subsubsection*{Results}
Tables~\ref{tab:few_shot_10_transposed_narrow2_clean} and \ref{tab:few_shot_5_transposed} show the few-shot results under the 10\% and 5\% settings. ST-Vision-LLM achieves the best performance on three of the four subsets in both settings. GATGPT is the only exception, ranking first on Trentino-Internet, while the strongest non-LLM baseline varies across datasets. ST-Vision-LLM also remains clearly ahead of GWNET and ResLSTM on Milan-Internet and Milan-SMS under both settings. These results indicate strong few-shot capability for data-scarce spatiotemporal prediction.

\subsection{Cross-Domain Few-Shot Fine-Tuning Prediction}

\begin{table}[h!]
\centering
\caption{Experimental results for cross-domain few-shot fine-tuning from the \textbf{Trentino-Call} source domain in terms of NRMSE. The input sequence length is set to $S=12$, and the prediction horizon is set to $K=36$. Models are fine-tuned using 2\% of the target domain's training data. The best result is marked in \textbf{bold} and the second best is \underline{underlined}.}
\label{tab:cross_domain_ft_transposed_optimized}
\begin{tabular}{lcccc}
\toprule
Method & \begin{tabular}[c]{@{}c@{}}Trentino- \\ Internet\end{tabular} & \begin{tabular}[c]{@{}c@{}}Trentino- \\ SMS\end{tabular} & \begin{tabular}[c]{@{}c@{}}Milan- \\ Internet\end{tabular} & \begin{tabular}[c]{@{}c@{}}Milan- \\ SMS\end{tabular} \\
\midrule
ST-Vision-LLM & \textbf{0.6638} & \textbf{0.9548} & \textbf{0.2150} & \textbf{0.7126} \\
ST-LLM & 1.0469 & 1.6283 & 0.5074 & \underline{0.9318} \\
GCNGPT & 0.9999 & 1.6218 & \underline{0.4819} & 0.9828 \\
GATGPT & 0.7790 & 1.4709 & 0.5071 & 0.9918 \\
Time-LLM & 0.9939 & 1.4384 & 0.5456 & 1.2201 \\
GWNET & \underline{0.6891} & 1.4040 & 0.8680 & 1.0182 \\
ResLSTM & 0.8566 & 1.5961 & 0.8253 & 1.1142 \\
STN & 1.0076 & \underline{1.3208} & 0.7980 & 0.9913 \\
\bottomrule
\end{tabular}
\end{table}

\subsubsection*{Setup}
LLMs are widely recognized for strong cross-domain learning and knowledge transfer capabilities \cite{ref_36}. To evaluate this capability, we follow the benchmark setup of the cross-domain prediction benchmark, pre-train all models on the complete Trentino-Call dataset, and then fine-tune them on Trentino-Internet, Trentino-SMS, Milan-Internet, and Milan-SMS using only 2\% of each target-domain training set before evaluation.

\subsubsection*{Results}
Table~\ref{tab:cross_domain_ft_transposed_optimized} shows that ST-Vision-LLM achieves the best performance on all four subsets, indicating stronger cross-domain adaptation than the baselines. Other LLM-based baselines are less consistent: for example, GATGPT ranks second on Trentino-Internet but not on the other datasets, while STN ranks second on Trentino-SMS but performs poorly elsewhere. A similar pattern is observed for the task-specific baselines, where GWNET is relatively competitive on Trentino-Internet but both GWNET and ResLSTM remain clearly behind ST-Vision-LLM on the Milan subsets.

\subsection{Zero-Shot Prediction}

\begin{table}[h!]
\centering
\caption{Experimental results for zero-shot prediction from the \textbf{Trentino-Call} source domain in terms of NRMSE. The input sequence length is set to $S=12$, and the prediction horizon is set to $K=36$. Models are evaluated directly on the target domain. The best result is marked in \textbf{bold} and the second best is \underline{underlined}.}
\label{tab:zero_shot_optimized}
\begin{tabular}{lcccc}
\toprule
Method & \begin{tabular}[c]{@{}c@{}}Trentino- \\ Internet\end{tabular} & \begin{tabular}[c]{@{}c@{}}Trentino- \\ SMS\end{tabular} & \begin{tabular}[c]{@{}c@{}}Milan- \\ Internet\end{tabular} & \begin{tabular}[c]{@{}c@{}}Milan- \\ SMS\end{tabular} \\
\midrule
ST-Vision-LLM & \textbf{1.0975} & \textbf{1.0541} & \textbf{0.4047} & \textbf{0.7560} \\
ST-LLM & 1.4578 & 2.2119 & 0.4940 & 1.2128 \\
GCNGPT & 1.2596 & 1.6848 & \underline{0.4384} & \underline{0.8171} \\
GATGPT & 1.4012 & 1.7934 & 0.4565 & 0.8515 \\
Time-LLM & \underline{1.1317} & \underline{1.5298} & 0.6253 & 1.3527 \\
GWNET & 1.3046 & 1.8726 & 0.7079 & 1.0983 \\
ResLSTM & 1.5969 & 2.2299 & 1.5463 & 1.5589 \\
STN & 1.2480 & 1.6420 & 0.8626 & 1.1715 \\
\bottomrule
\end{tabular}
\end{table}

\subsubsection*{Setup}
LLMs also have great potential as effective zero-shot learners \cite{ref_29}. In this setup, we evaluate the zero-shot learning capability of ST-Vision-LLM within a cross-domain transfer framework. We fully train a model on one dataset and directly evaluate its performance on another, during which the model is not exposed to any data samples from the target domain. We pre-train the models on Trentino-Call and then directly evaluate them on the Trentino-Internet, Trentino-SMS, Milan-Internet, and Milan-SMS datasets, with other experimental settings remaining consistent with those in the cross-domain prediction benchmark.

\subsubsection*{Results}
Table~\ref{tab:zero_shot_optimized} shows that ST-Vision-LLM achieves the best performance on all four target domains. The second-best model varies across datasets, but ST-Vision-LLM remains consistently ahead. On Milan-Internet, for example, it reduces the NRMSE by about 7.7\% compared with GCNGPT and by more than 18\% compared with ST-LLM. Other baselines, including STN, GWNET, and ResLSTM, degrade more substantially in the zero-shot setting. These results indicate strong zero-shot transfer capability.

\subsection{Ablation Study}

\begin{table}[h!]
\centering
\caption{Ablation study on Milan-Internet in terms of MAE, RMSE, NRMSE, and output length. The input sequence length is set to $S=12$, and the prediction horizon is set to $K=10$.}
\label{tab:ablation_study}
\begin{tabular}{lcccc}
\toprule
Method & MAE & RMSE & NRMSE & Out. Tokens \\
\midrule
ST-Vision-LLM & 27.49 & 41.85 & 0.134 & 13 \\
w/o GRPO & 29.12 & 44.35 & 0.142 & 13 \\
w/o Image Encoder & 32.92 & 47.86 & 0.153 & 13 \\
w/o LLM Backbone & 148.45 & 218.19 & 0.697 & 13 \\
w/o Num. Enc. + Dec. & 26.70 & 39.74 & 0.127 & 112 \\
w/o Num. Enc. + Int. & 28.15 & 40.80 & 0.130 & 42 \\
\bottomrule
\end{tabular}
\end{table}

Table~\ref{tab:ablation_study} presents the ablation results of the main components of ST-Vision-LLM. Removing either the image encoder or the LLM backbone leads to a clear degradation in prediction accuracy. Specifically, after removing the image encoder, the MAE, RMSE, and NRMSE all increase noticeably, indicating that the global visual spatial context is effective for the forecasting task. After removing the LLM backbone, the performance degrades much more severely, showing that the sequence modeling and generation capability provided by the language-model backbone is indispensable to the overall framework.

Removing the GRPO stage also leads to a consistent performance drop. Compared with the full model, the variant without GRPO yields higher MAE, RMSE, and NRMSE, which indicates that the second-stage optimization further improves prediction accuracy on top of the SFT model.

The comparison among different value representations reveals a clear trade-off between predictive performance and output length. In Table~\ref{tab:ablation_study}, ``Num. Enc.'' denotes Numerical Encoding, ``Dec.'' denotes Decimal String, and ``Int.'' denotes Integer Approximation. The two variants without Numerical Encoding directly generate either decimal strings or integer-approximated strings. The Decimal String setting achieves the lowest error in this table, while the Integer Approximation setting also remains competitive. However, their output lengths increase to 112 and 42 tokens, respectively, whereas the full model with Numerical Encoding requires only 13 output tokens. We additionally report Output Tokens because the output stage of an LLM is autoregressive and the decoding latency is directly related to the number of generated tokens, while the input length mainly affects the prefill stage, which is largely parallelizable on GPUs. These results indicate that Numerical Encoding provides a more balanced trade-off between predictive performance and generation efficiency.

Additional numerical-encoding efficiency analysis, qualitative visualization, inference-efficiency results, and spatial-region stability analysis are provided in the appendix.

\section{Conclusion}
ST-Vision-LLM reframes spatiotemporal traffic forecasting as a vision-language problem, encoding historical data as visual patches for global context and using textual prompts for targeted predictions. A core innovation is our numerical encoding scheme, combined with a two-stage alignment fine-tuning, which endows the LLM with essential numeracy. The framework achieves state-of-the-art performance, demonstrating exceptional generalization and data efficiency, especially in few-shot, cross-domain, and zero-shot scenarios. This validates our hypothesis that leveraging the global context of Vision-LLM overcomes the limitations of traditional models, establishing a powerful and generalizable new paradigm for spatiotemporal prediction.

\clearpage
\appendix
\subsection{Numerical Encoding Efficiency Analysis}
To further quantify the efficiency benefit of Numerical Encoding, we report three complementary statistics. First, when floating-point values are sampled from the range $[0,10000]$ and represented with six decimal places, a standard decimal string requires 10.89 tokens on average, whereas Numerical Encoding uses exactly one token per value, reducing the average by 9.89 tokens (90.81\%). Second, under the commonly used setting of $S=12$ and $K=36$, the numerical sequence itself is reduced from 614.69 tokens to 48.00 tokens on average, while the full context is reduced from 1034.57 tokens to 467.87 tokens on average, corresponding to a 54.77\% reduction. Here, the full-context count includes image tokens, text instructions, the numerical sequence, and the end-of-sequence (EOS) token. Third, under the same setting, the output sequence length is reduced from 465.01 tokens to 39.00 tokens on average, corresponding to a reduction of 426.01 tokens (91.61\%).

We also report the additional cost of the two-stage numerical alignment fine-tuning. The Stage 1 training set contains 799,600 samples and 36.78M tokens. The Stage 2 training set contains 108,750 samples and 17.80M tokens on average, and the validation set contains 5,444 samples and 0.89M tokens on average. The two stages together therefore involve 54.58M training tokens on average. On a single RTX 5060 GPU, the two-stage numerical alignment training takes about 6 hours in total. We consider this additional cost acceptable because it yields substantial gains in both context compression and output-length compression.

\subsection{Inference Efficiency and Deployment Discussion}
\begin{table}[h!]
\centering
\caption{Inference latency and output length on Milan-Internet. The setting is $S=12$ and $K=10$.}
\label{tab:inference_efficiency}
\begin{tabular}{lccc}
\toprule
Method & \begin{tabular}[c]{@{}c@{}}Single-Cell \\ Latency (s)\end{tabular} & \begin{tabular}[c]{@{}c@{}}$10\times10$ Region \\ Latency (s)\end{tabular} & \begin{tabular}[c]{@{}c@{}}Output \\ Tokens\end{tabular} \\
\midrule
ST-Vision-LLM & 0.41 & 1.95 & 13 \\
w/o Num. Enc. + Dec. & 2.13 & 14.15 & 112 \\
\bottomrule
\end{tabular}
\end{table}

\subsubsection*{Setup}
We further evaluate the inference efficiency of ST-Vision-LLM on Milan-Internet under the setting of $S=12$ and $K=10$. The test region is $(x,y)\in[45,55)\times[45,55)$, and all measurements are conducted on a single NVIDIA 5090 GPU with 32GB memory. We report both the average latency for single-cell prediction and the total latency for point-by-point prediction over the full $10\times10$ region.

\subsubsection*{Results}
Table~\ref{tab:inference_efficiency} reports average latency and average output length, where ``Num. Enc.'' denotes Numerical Encoding and ``Dec.'' denotes Decimal String. ST-Vision-LLM requires only 0.41 seconds for single-cell prediction and 1.95 seconds for the full $10\times10$ region, while the variant w/o Num. Enc. + Dec. requires 2.13 and 14.15 seconds, respectively. The corresponding output length is reduced from 112 tokens to 13 tokens. Since the decoding stage of an LLM is autoregressive, this output-length reduction directly translates into lower inference latency. Moreover, the region-level latency of 1.95 seconds remains far below the 10-minute observation interval of the dataset, indicating that the framework is compatible with real-time deployment in the current setting.

For the region-level measurement, the image context and the shared prompt prefix are reused through a shared-prefix key-value (KV) cache. As a result, the reported region latency reflects the actual cost of predicting multiple locations under the same global historical context, rather than repeatedly recomputing the identical visual prefix for each target point.

\subsection{Spatial Region Stability Analysis}

\begin{table}[h!]
\centering
\caption{Performance across different spatial regions on Milan-Internet. The input sequence length is set to $S=12$, and the prediction horizon is set to $K=10$.}
\label{tab:region_stability}
\begin{tabular}{lccc}
\toprule
Region & MAE & RMSE & NRMSE \\
\midrule
Center Region & 27.49 & 41.85 & 0.134 \\
High1 & 10.59 & 19.43 & 0.153 \\
High2 & 13.68 & 21.30 & 0.148 \\
High3 & 29.36 & 67.06 & 0.157 \\
Low1 & 4.83 & 7.53 & 0.147 \\
Low2 & 5.83 & 10.05 & 0.177 \\
\bottomrule
\end{tabular}
\end{table}

\subsubsection*{Setup}
To examine whether the empirical results are sensitive to the choice of the evaluation region, we further test ST-Vision-LLM on several additional spatial regions of the Milan-Internet dataset under the same setting as the main benchmark, namely using 12 historical frames to predict the next 10 frames. The evaluated regions are Center Region $(x,y)\in[45,55)\times[45,55)$, High1 $(x,y)\in[30,40)\times[50,60)$, High2 $(x,y)\in[80,90)\times[70,80)$, High3 $(x,y)\in[50,60)\times[60,70)$, Low1 $(x,y)\in[20,30)\times[80,90)$, and Low2 $(x,y)\in[70,80)\times[30,40)$.

\subsubsection*{Results}
Table~\ref{tab:region_stability} reports the performance on the central region together with several additional high-traffic and low-traffic regions. Here, High1--High3 denote regions with relatively high traffic intensity, while Low1--Low2 denote regions with relatively low traffic intensity. As expected, the absolute traffic scale affects the magnitude of MAE and RMSE, so these two metrics are not directly comparable across regions with very different traffic levels.

For this reason, NRMSE provides a more suitable criterion for cross-region comparison. The results show that the NRMSE values remain within a relatively close range across all tested regions, without an obvious instability when moving from the central region to higher-traffic or lower-traffic areas. This suggests that the effectiveness of ST-Vision-LLM does not rely on one specific spatial location and remains reasonably stable under spatial region variation.

\subsection{Prediction Visualization}

\begin{figure}[h!]
\centering
\includegraphics[width=\columnwidth]{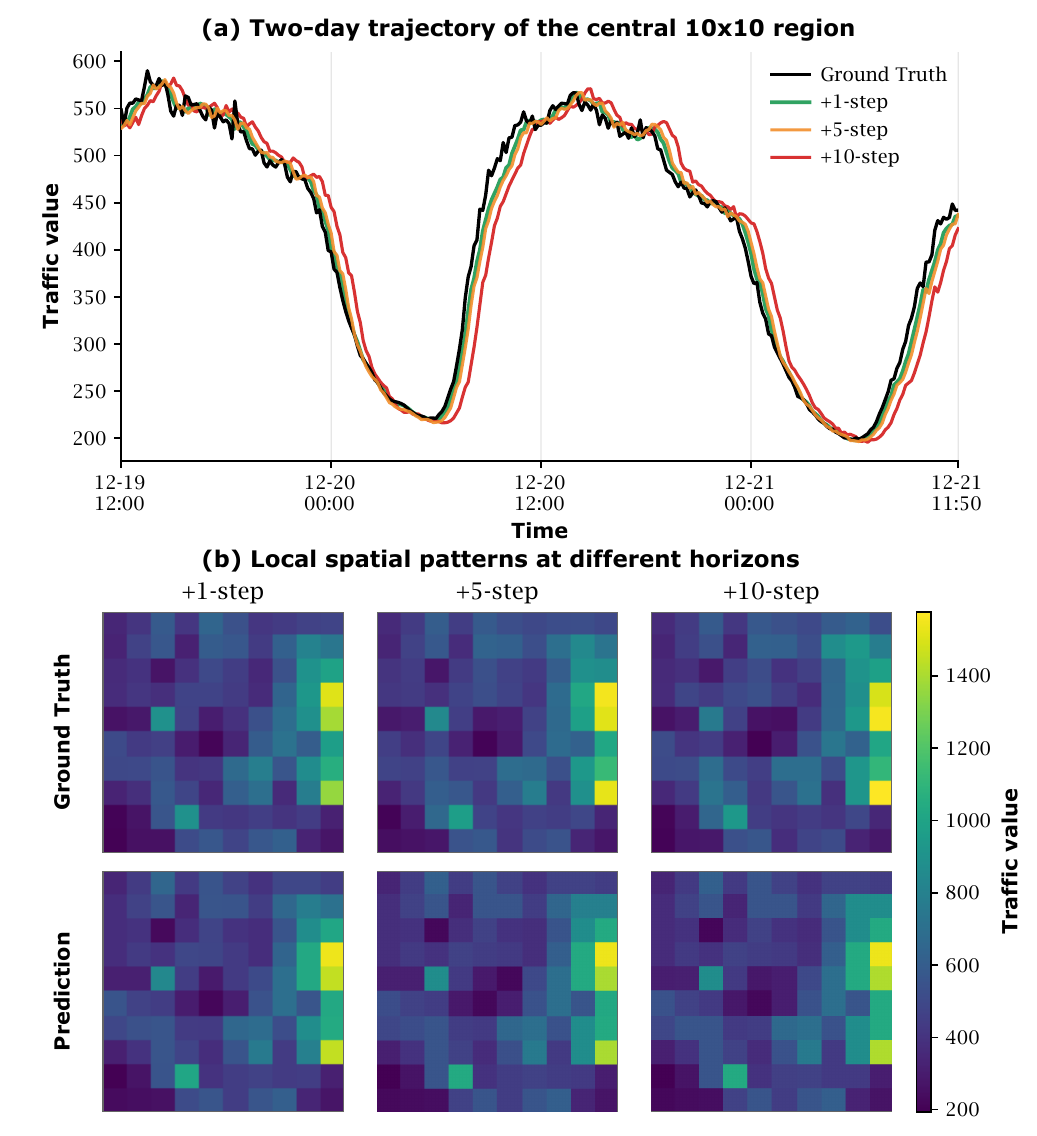}
\caption{Visualization results on Milan-Internet under the setting of using the past 12 time points to predict the next 10 time points. Panel (a) shows the two-day average traffic trajectory over the central $10\times10$ region, together with the rolling predictions at the $+1$-, $+5$-, and $+10$-step horizons. Panel (b) shows the corresponding local spatial patterns at the three prediction horizons for the sample starting at 2013-12-20 10:00:00, where the first row is the ground truth and the second row is the prediction. All heatmaps share the same spatial window and color scale.}
\label{fig:prediction_visualization}
\end{figure}

\subsubsection*{Setup}
We further present a visualization example on Milan-Internet under the setting of using the past 12 time points to predict the next 10 time points. Figure~\ref{fig:prediction_visualization}(a) reports the average traffic trajectory over the central $10\times10$ region across two consecutive days, together with the rolling predictions at the $+1$-, $+5$-, and $+10$-step horizons. Figure~\ref{fig:prediction_visualization}(b) shows the corresponding local spatial patterns for the sample starting at 2013-12-20 10:00:00.

\subsubsection*{Results}
Figure~\ref{fig:prediction_visualization}(a) shows that the predicted trajectories follow the overall rise-and-fall trend of the ground-truth traffic well. As expected, the prediction at the shorter horizon remains closer to the ground truth, while the deviation gradually increases from the $+1$-step horizon to the $+10$-step horizon. Nevertheless, even at the longer horizon, the model still captures the major temporal trend consistently.

Figure~\ref{fig:prediction_visualization}(b) further shows that the predicted heatmaps preserve the main local spatial patterns across the $+1$-, $+5$-, and $+10$-step horizons. In particular, the major high-value and low-value regions, as well as the dominant spatial gradients, remain visually consistent with the ground truth. This indicates that the model not only tracks temporal variation effectively, but also maintains reasonable local spatial structure under different prediction horizons.

\subsection{Instruction Template}\label{app:prompt-template}
Full instruction template used in training and inference is shown below. The historical global traffic context is composed of \textit{S} input images, and the text prompt contains the task instruction, data type, time range, target coordinates, normalization parameters, and the target point's historical numerical sequence.

\vspace{0.5em}
\noindent\fbox{%
\parbox{0.96\columnwidth}{%
\raggedright\ttfamily\small
<image> repeated \textit{S} times\par
Predict the traffic data of the next \textit{output\_len} frames based on the past \textit{input\_len} frames of traffic data.\par
The input traffic data is a two-dimensional grid (\textit{grid\_h}, \textit{grid\_w}) that has already been embedded in the above images context.\par
Predict the traffic data at one specific point, with coordinates (\textit{x}, \textit{y}).\par
Data type: \textit{data\_type}\par
Input Time range: \textit{time\_start} -> \textit{time\_end}\par
Coordinates of the data point to predict: (\textit{x}, \textit{y}).\par
Normalization parameters of the input traffic data grid: max\_num = \textit{tmax}, p = \textit{p}\par
Historical data of the prediction point (in chronological order):\par
[<|FP...|><|FP...|>...<|FP...|>]\par
Output:\par
[<|FP...|><|FP...|>...<|FP...|>]
}}

\vspace{0.5em}
\noindent Here, \textit{S} denotes the number of historical traffic images; \textit{input\_len} and \textit{output\_len} denote the input and output sequence lengths; \textit{grid\_h} and \textit{grid\_w} denote the spatial height and width of the traffic grid; \textit{x} and \textit{y} denote the target coordinates; \textit{data\_type} denotes the traffic modality; \textit{time\_start} and \textit{time\_end} denote the start and end timestamps of the current input window; \textit{tmax} denotes the maximum traffic value used in normalization; and \textit{p} denotes the normalization hyperparameter. All italicized items in the template are placeholders to be replaced with instance-specific values, rather than literal strings in the final prompt.

\end{document}